\documentclass{article}

\usepackage[final]{neurips_2019}

\usepackage[utf8]{inputenc} 
\usepackage[T1]{fontenc}    
\usepackage{hyperref}       
\usepackage{url}            
\usepackage{booktabs}       
\usepackage{amsfonts}       
\usepackage{nicefrac}       
\usepackage{microtype}      

\usepackage{graphicx}
\usepackage{subcaption}
\usepackage{mathrsfs}
\usepackage{xcolor}
\usepackage{multirow}
\usepackage{multicol}
\usepackage{MnSymbol}

\definecolor{darkgreen}{RGB}{0,100,0}
\definecolor{Gray}{gray}{0.9}

\graphicspath{{img/}{./}}

\usepackage{amsmath}

\usepackage[ruled]{algorithm2e}
\makeatletter
\newcommand{\algorithmstyle}[1]{\renewcommand{\algocf@style}{#1}}
\newcommand{\removelatexerror}{\let\@latex@error\@gobble}
\makeatother

\title{Reducing Selection Bias in Counterfactual Reasoning for Individual Treatment Effects Estimation}

\author{%
  Zichen Zhang$^{\dagger}$, Qingfeng Lan$^{\dagger}$, Lei Ding$^{\ddagger}$, Yue Wang$^{\ddagger}$, Negar Hassanpour$^{\dagger}$, Russell Greiner$^{\dagger}$\\
  $^{\dagger}$Department of Computing Science\\
  $^{\ddagger}$Department of Mathematical and Statistical Sciences\\
  University of Alberta\\
  \texttt{ \{zichen2,qlan3,lding1,yue9,hassanpo,rgreiner\}@ualberta.ca } \\
}

\begin{document}

\maketitle

\begin{abstract}
  Counterfactual reasoning is an important paradigm applicable in many fields, such as healthcare, economics, and education. In this work, we propose a novel method to address the issue of \textit{selection bias}. We learn two groups of latent random variables, where one group corresponds to variables that only cause selection bias, and the other group is relevant for outcome prediction. They are learned by an auto-encoder where an additional regularized loss based on Pearson Correlation Coefficient (PCC) encourages the de-correlation between the two groups of random variables. 
  This allows for explicitly alleviating selection bias by only keeping the latent variables that are relevant for estimating individual treatment effects. Experimental results on a synthetic toy dataset and a benchmark dataset show that our algorithm is able to achieve state-of-the-art performance and improve the result of its counterpart that does not explicitly model the selection bias.
\end{abstract}

\section{Introduction}
\label{intro}
Studying the causal effect of different treatments on individuals to assist in decision making is an essential problem in various fields. Examples include a doctor deciding the most effective medical treatment for a specific patient, a company deciding the most profitable commercial advertisement for a specific product, etc.

In this paper, we focus on understanding individual-level causal effects in healthcare. 
Access of many observational data in this field allows us to develop methods for predicting individual-level causal effects. However, many challenges remain. 
The first challenge is that the observational data at the individual level tells us only the outcome of received treatments (the factuals), whereas the responses of the alternative treatments (the counterfactuals) are never available. 
For example, if a patient is given surgery, we would not be able to observe the true effect of applying medication instead.
This setting is called \textit{counterfactual reasoning}, that is, to predict the individual treatment effect of the counterfactual treatment. 
The second challenge is that the data often exhibits \textit{selection bias}~\citep{imbens2015causal}.
For example, patient living in the rural area may not have access to a certain medication. 
Consequently, there are only a few, if any, patients receiving that medication in the dataset, i.e., the observational data have a \textit{selection bias}. In this case, the home address affects only the treatment.
In other cases, there are factors that affect both the treatment and the respective outcome, called \textit{confounder}.
It also causes the difficulty of predicting the causal effects since it partially leads to the \textit{selection bias}. 
For example, 
it is more likely for a doctor to prescribe surgery to younger patients while to give medication to older patients. 
On the other hand, ages may affect the potential outcome regardless of the treatment given. 

We represent the observed features of each patient, such as age and gender, as a random vector $\mathbf{X}$. 
For simplicity, we assume that there are only two treatments, denoted as a binary variable $T \in \{0, 1\}$. The patients who receive treatment $T=0$ or $T=1$ are in the control group and treatment group, respectively. Consequently, there are two possible outcomes $Y^0$ and $Y^1$ corresponding to each treatment option: $T=0$ and $T=1$. However, we do not have access to both of the outcomes. For each patient, we only observe the outcome corresponding to the received treatment. We denote all the observed outcomes (the factuals) as $Y^f$ and all the unobserved outcomes (the counterfactuals) as $Y^{cf}$. Moreover, the selection bias can be expressed as $p(T|\mathbf{X}) \neq p(T)$.
The goal is to estimate the Individual Treatment Effect (ITE), i.e.,  $E[Y^1-Y^0]$ for each individual.

\begin{figure}
\begin{center}
    \includegraphics[width=0.3\textwidth]{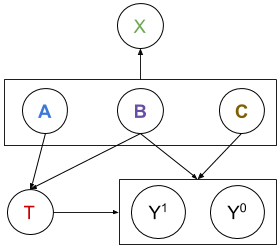}
    \caption{The proposed causal graph for individual treatment effect estimation}
    \label{fig:graphic_model}
\end{center}
\end{figure}

The causal graph in our analysis is shown in  Figure~\ref{fig:graphic_model}, inspired by \citet{CFR-ISW}.
We assume that the covariate $\mathbf{X}$ is generated by three types of latent variables. The first type $\mathbf{A}$ includes the latent variables that only affect treatment selection procedure but do not determine outcomes. Type $\mathbf{B}$ are the confounders which influence both treatments and outcomes. The last type $\mathbf{C}$ only affects outcomes.

Our main contribution is that we propose a novel method that separates the learned feature representations into two parts, corresponding to $\mathbf{A}$ and $\mathbf{BC}$ described above. Then we reduce the selection bias by using only the representation of type $\mathbf{BC}$ to predict the outcomes. We test our algorithm on two datasets: a synthetic toy dataset and a benchmark dataset simulated from real-world data. The results show that our method helps to improve the prediction performance in many settings.

\textbf{Assumptions}
Similar to the work by \citet{shalit17a}, we assume that there exists a joint distribution $p(\mathbf{X},T,Y^0,Y^1)$ with ``strong ignorability'' assumption: $Y \upmodels T | \mathbf{X}$ and $0<p(t = 1|x)<1, \forall x \in \mathcal{X}$. This is sufficient for the ITE to be identifiable~\citep{imbens2009recent}.
We also assume that the outcomes of the samples $(\mathbf{x}_1,t_1,y_1),...(\mathbf{x}_n,t_n,y_n)$ are generated from $y^i \sim p(y^{t_i}|\mathbf{x}_i)$. 

\section{Related works}
\label{related}
In the work of \citet{johansson2016learning}, ideas from domain adaptation and representation learning were combined. Besides learning a latent representation for outcome prediction, the discrepancy distance~\citep{mansour2009domain} was introduced as a distribution distance metric. By minimizing the discrepancy distance, the distributions of populations with different treatments were balanced to reduce the selection bias. 
On the network structure, they simply concatenated the treatment $t$ with the representation $\Phi$ which easily led to information loss of treatment $t$.

\citet{shalit17a} improved this line of work and proved a generalization error-bound for estimating ITE.
The improvements they made were two-fold. First, they introduced a branching network structure where one branch of the network learned the prediction of treated outcome ($t=1$) and the other branch learned the prediction of the outcome under control ($t=0$). 
This new network structure solved the issue in the work of \citet{johansson2016learning} of losing the influence of $t$ when the dimension of the representation was large.
Second, they introduced a measure of distance between two distributions $p(\mathbf{x}|t=1)$ and $p(\mathbf{x}|t=0)$, called Integral Probability Metric (IPM). They showed that the expected error of the ITE prediction was upper bounded by the error of learning $Y^1$ and $Y^0$, plus the IPM term.
This IPM measure was therefore used in the loss function to encourage that the two distributions of representations being closer.

From the perspective of network structure, our work is closely related to \citet{atan2018deep}. In this work, a latent representation is learned by using an auto-encoder. By jointly minimizing the reconstruction loss and the distribution distance between different representation groups, it balanced between information loss and bias reduction. 
However, the entire learned representation was then used for outcome prediction. This could inevitably contain features that is not useful for outcome prediction therefore counteracting the effect of bias reduction in the first step.
We address this issue by learning the features that only causes selection bias and discard them during the outcome prediction.

In terms of disentangling two categories of representations, we are inspired by the work from \citet{cheung_discovering_2014}. This work was in the domain of image classification, where they learned the features of class-independent variations $Z$ apart from the features for classification. They introduced a cross-covariance penalty (XCov) for this purpose. It disentangled factors like the hand-writing style from the digits labels. We improve their work by introducing a penalty term that better reflects the correlation between random variables, as detailed in the next section.

\section{Proposed method}
\label{sec:proposed}

\subsection{Network Architecture}

%
Following the model of deep-treat in \citep{atan2018deep} and the Counter Factual Regression (CFR) framework in \citep{shalit17a}, the overall structure of our method is an autoencoder where the representation learning stage is also followed by an outcome prediction stage that branches based on the treatment $t$ of the input sample $\mathbf{x}$.

In order to reduce the selection bias, we explicitly model the bias (latent variables type \textbf{A}, note that it is in \textbf{bold} denoting a vector of random variables) in the learned representation $\Phi(\mathbf{x})$ and separate it from the rest of the features (latent variables type \textbf{BC}) that are relevant for the outcome prediction.
Since the bias variables \textbf{A} do not play a part in the outcome prediction, only the variables \textbf{BC} are then used as the input to the downstream prediction network to predict the outcomes for various treatments $\hat{y}^1(\mathbf{x})$ and $\hat{y}^0(\mathbf{x})$ (we consider binary treatment in this work, i.e, $t=1$ or $t=0$).

\begin{figure}[tbp]
\begin{center}
    \includegraphics[scale=0.6]{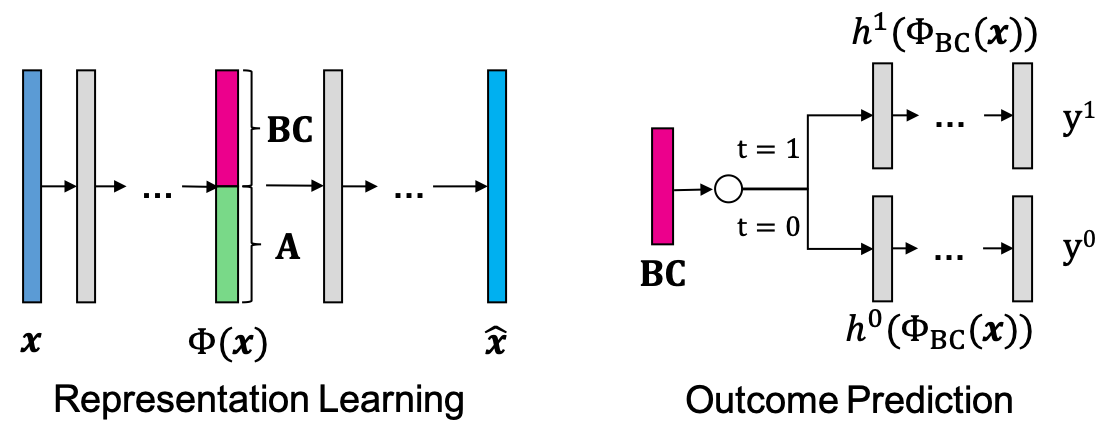}
    \caption{The architecture of the proposed model RSB-Net. The representation learning network is an auto-encoder that learns the bias variables \textbf{A} and the variables \textbf{BC} that are relevant for prediction. The outcome prediction network has a branching structure predicting the outcome $y^t$ based on the treatment $t$ and the representation of \textbf{BC}.} 
    \vspace{-5pt}
    \label{fig:model}
\end{center}
\end{figure}

The architecture of the proposed method, named RSB-Net (stands for Reducing Selection Bias), is illustrated in Fig.~\ref{fig:model}. In the next subsection, we explain how this network can be trained to explicitly learn the two groups of latent variables \textbf{A} and \textbf{BC}.

\subsection{Loss Function}
On the high level, the proposed neural network is trained end-to-end with a hybrid loss that is a weighted sum of the following: a prediction loss $L_\text{pred}$, a distribution loss $L_\text{IPM}$, an input reconstruction loss $L_\text{recons}$ and a loss based on Pearson correlation coefficient $L_\text{pcc}$.
\begin{equation}
    \mathcal{L} = L_\text{pred} + \alpha L_\text{IPM} + \beta L_\text{recon} + \gamma L_\text{pcc} + \lambda R(W) 
    \label{eq:loss}
\end{equation}
where $R(W)$ is the regularization term of the network weights $W$ that penalizes complexity of the network, and $\alpha$, $\beta$, $\gamma$, $\lambda$ are the weights of the loss terms.

\textbf{Prediction Loss and Distribution Loss} 
This is the supervised loss proposed in~\citet{shalit17a}. 
For a batch of data samples ${\mathbf{x}_i,t_i}$, we aim to predict the factual outcome $\hat{y}_i^{t_i}$. The prediction loss is defined on the factual outcome $y^{t_i}$ using a weighted squared loss as
\begin{equation}
    L_{pred}= \frac{1}{N} w_i\|h^{t_i}(\Phi_{\mathrm{BC}}(\mathbf{x_i})) - y_i^{t_i}\|_{2}^{2}    
\end{equation}
where $N$ is the sample size, $w_i=\frac{t_i}{2u}+\frac{1-t_i}{2(1-u)}$, $u=\frac{1}{N}\sum_{i}t_{i}$. 
Note that $u$ is the probability of choosing treatment $t=1$ in the entire population, i.e. $u=p(t=1)$. $w_i$ compensates for the size difference in different treatment arms. 
The distribution loss, using Integral Probability Metric(IPM) is defined as  \begin{equation}
    L_\text{IPM}= \text{IPM} (\Phi_\mathrm{BC}(\mathbf{x}_i)_{i:t_i=0}, \Phi_\mathrm{BC}(\mathbf{x}_i)_{i:t_i=1}) 
 \end{equation}
It measures the distribution distance of the latent representation of the treated group and the control group, i.e. between $\Phi(\mathbf{x}|t=1)$ and $\Phi(\mathbf{x}|t=0)$.

\textbf{Reconstruction Loss} 
Inspired by the approach in \cite{cheung_discovering_2014} and \cite{atan2018deep}, we use an auto-encoder with a squared $L_2$ loss to learn a set of latent representation for both groups of random variables \textbf{A} and \textbf{BC}. The loss is defined as
$L_\text{recons} = \| \hat{\mathbf{x}} - \mathbf{x}\|_{2}^{2}$, 
where $\hat{\mathbf{x}}$ is the reconstruction of the input features of the sample $\mathbf{x}$.

\textbf{PCC Loss} 
To explicitly learn the random variables \textbf{A} and \textbf{BC},
the latent representation $\Phi(\mathbf{x})$ is first split into two parts (the ratio is a hyperparameter), denoted as $\Phi_{\mathbf{A}}(\mathbf{x}) \in \mathbb{R}^m$ and $\Phi_{\mathbf{BC}}(\mathbf{x}) \in \mathbb{R}^n$, corresponding to \textbf{A} and \textbf{BC} respectively. $m$ and $n$ denote the dimension of the vectors for each sample $\mathbf{x}$.
We would like $\Phi_{\mathbf{A}}(\mathbf{x})$ and $\Phi_{\mathbf{BC}}(\mathbf{x})$ to be de-correlated in the learned representation. 

To this end, we define a loss based on Pearson correlation coefficient (PCC):
\begin{equation}
    L_\text{pcc} = 
    \frac{1}{2mn}\sum^{m}_{i=1}\sum^{n}_{j=1}[ \frac{\frac{1}{N}\sum_{k=1}^{N}(\Phi_{\mathrm{A}}(\mathbf{x}_k)_{i}-\overline{\Phi_{\mathrm{A}}}_{i})(\Phi_{\mathbf{BC}}(\mathbf{x}_k)_{j}-\overline{\Phi_{\mathbf{BC}}}_{j})}{\sigma({\Phi_{\mathrm{A}_i}}) \sigma({\Phi_{\mathrm{BC}_j}})} ]^{2} 
    \label{eq:pcc}
\end{equation}
where $\Phi_{\mathbf{A}}(\mathbf{x}_k)_{i}$ is the $i$-th element of vector $\Phi_{\mathbf{A}}(\mathbf{x}_k)$ for sample $k$, $\overline{\Phi_{\mathbf{A}}}_{i}$ is the mean value of the $i$-th element of vector $\Phi_{\mathbf{A}}$ for all samples. We use similar notations for $\Phi_{\mathbf{BC}}$.
The idea of this loss is to take the mean of the squared PCC between every pair of random variables formed by one entry in vector $\Phi_{\mathbf{A}}$ and one entry $\Phi_{\mathbf{BC}}$.
Since PCC $\in [-1,1]$, we have the range of this loss $L_\text{pcc} \in [0, 0.5]$, reaching the minimum when features $\Phi_{\mathbf{A}}$ and $\Phi_{\mathbf{BC}}$ are linearly independent.

\begin{algorithm}[H]
\SetNoFillComment
 \KwIn{Factual samples \{$(\mathbf{x}_1,t_1,y^f_1)$,...,$(\mathbf{x}_N,t_N,y^f_N)$\}, coefficients of the loss terms: $\alpha$, $\beta$, $\gamma$, $\lambda$\ 
 minibatch size $m$}
 Compute $u=\frac{1}{N}\sum_{i=1}^{N}t_{i}$ and the sample weight: $w_{i}=\frac{t_{i}}{2u}+\frac{1-t_{i}}{2(1-u)}$ for $i=1,\cdots,N$\;
 Initialize the weights $\mathbf{W}$ in neural networks\;
  \Repeat{max iterations}
  {
    Sample a mini-batch $\{i_1,i_2, \cdots ,i_m \} \subseteq \{1,2,...,N\}$\;
    \ForEach{sample $(\mathbf{x},y,t)$}
    {
      \tcc{Representation Learning}
      Compute $\Phi(\mathbf{x})$\ and the reconstructed $\mathbf{x}$: ${\hat{\mathbf{x}}=\Psi(\Phi(\mathbf{x}))}$\;
      \tcc{Outcome Prediction}
      Split the representation $\Phi(\mathbf{x})$ into bias $\Phi_\mathrm{A}(\mathbf{x})$ and representation $\Phi_\mathrm{BC}(\mathbf{x})$ \;
      Compute the predicted outcome corresponding to treatment $t$, $y^t=h^t(\Phi_{\mathrm{BC}}(\mathbf{x}))$;
    }
    Compute the prediction loss $ L_\text{pred} = \frac{1}{m} \sum_{j=1}^{m} w_{i_j} (\ell_2(\hat{\mathbf{y}}^{t_{i_j}}_{i_j}, y^{t_{i_j}}_{i_j}))^2$ \;
    Compute the distribution loss $L_\text{IPM} = \ell_\mathrm{IPM}(\Phi_\mathrm{BC}(\mathbf{x}_i)_{i:t_i=0}, \Phi_\mathrm{BC}(\mathbf{x}_i)_{i:t_i=1})$ \;
    Compute the reconstruction loss $ L_\text{recon} = \frac{1}{m} \sum_{j=1}^{m} (\ell_2(\hat{\mathbf{x}}_{i_j}, \mathbf{x}_{i_j}))^2$ \;
    Compute the Pearson correlation coefficient loss $L_\text{pcc}$ defined in Eq.~\ref{eq:pcc}\;
    Sum up the above loss functions and add regularization $R$ to get the total loss f
    $$\mathcal{L} = L_\text{pred} + \alpha L_\text{IPM} + \beta L_\text{recon} + \gamma L_\text{pcc} + \lambda R(\mathbf{W})$$ 
    Optimize all weights $\mathbf{W}$ in the neural networks\;
    }
    \KwOut{Neural network weights $\mathbf{W}$}
 \caption{RSB-Net}
 \label{alg:RSB}
\end{algorithm}

\section{Experiments}
\label{sec:Experiments}
In this section, we present the experimental results of our proposed method on a new synthetic toy dataset and a benchmark dataset.
We compare our method with the following baseline methods:
k-nearest neighbor (kNN),
Bayesian Additive Regressoin Trees (BART)~\citep{chipman2010bart},
Balancing Neural Network (BNN)~\citep{johansson2016learning},
Deep-Treat~\citep{atan2018deep},
Treatment-Agnostic Representation Network (TARNET)~
\citep{shalit17a},
Counterfactual Regression with Wasserstein metric (CFRW)~\citep{shalit17a},
Counterfactual Regression with Importance Sampling Weights (CFR-ISW)~
\citep{CFR-ISW}, Causal Effect Variational
Autoencoder (CEVAE)~\citep{louizos2017causal}.

\subsection{Evaluation Metrics}
\label{sec:metric}
Our goal is to estimate the Individual Treatment Effects (ITE), which measures the difference between possible outcomes for each patient. The ground truth ITE and estimated ITE are defined as follows: \[\tau(\mathbf{x}):=\mathbb{E}[Y^1 - Y^0|\mathbf{x}] \quad \text{and} \quad \hat{\tau}(\mathbf{x}) =  h^1(\Phi(\mathbf{x})) -  h^0(\Phi(\mathbf{x}))\]
where $\Phi(\mathbf{x})$ is the representation function of the form $\Phi:\mathcal{X} \rightarrow \mathcal{R}$ that transforms $\mathbf{x}$ from the sample space $\mathcal{X}$ into the learned representation space $\mathcal{R}$, and $h$ is the hypothesis function $\mathcal{R} \times \{0, 1\} \rightarrow \mathcal{Y} $, defined over the representation space $\mathcal{R}$ and the treatment $t\in{\{0,1\}}$, mapping to the output space $\mathcal{Y}$.

Following the setup in~\citep{shalit17a}, we use the noiseless outcomes $\mu^1$ and $\mu^0$ as the ground truth so that 
$\tau(\mathbf{x}) = \mu^1(\mathbf{x}) - \mu^0(\mathbf{x})$.
And we use two metrics to evaluate estimated ITE. The first one is Precision in Estimation of Heterogeneous Effect (PEHE) defined as: \[\epsilon_\text{PEHE} = \frac{1}{N} \sum_i^N (\hat{\tau}(\mathbf{x}_i) - {\tau}(\mathbf{x}_i))^2 \] where $N$ is the sample size. This measures the mean squared difference between the estimated ITE and true ITE. Note that $\epsilon_\text{PEHE}$ is originally defined on a continuous distribution~\citep{hill_bayesian_2011}. Here we use the discrete version for finite samples. 

Another metric is the bias of the Average Treatment Effect (ATE):
\[\epsilon_\text{ATE} = | \widehat{\text{ATE}} - \text{ATE} | = | \frac{1}{N} \sum_{i=1}^N (\hat{\tau}(\mathbf{x}_i) - {\tau}(\mathbf{x}_i)) | \]
where $\text{ATE} = \mathbb{E}[{\tau}(\mathbf{x}_i)] = \frac{1}{N} \sum_{i=1}^N{\tau}(\mathbf{x}_i)$.
This measures the population difference between the expectation of the estimated ITE and true ITE.

For all experiments, we report the within-sample and out-of-sample mean and standard errors of $\sqrt{\epsilon_\text{PEHE}}$ and $\epsilon_\text{ATE}$ following the literature.
Within-sample takes into account the entire training data, including the training and validation split. Out-of-sample result measures the performance on the hold-out test dataset.

\subsection{Implementation details}
\label{implementation}
We implemented our RSB net using TensorFlow, based on the code\footnote{\url{https://github.com/clinicalml/cfrnet}} provided by~\citet{johansson2016learning}. 


Preprocessing of the data have not been commonly used or mentioned in the literature.
In the hyper-parameter tuning, we tested different preprocessing methods such as Z-score Standardization and min-max Normalization. Our empirical result across both datasets suggests that min-max Normalization either improves or shows no impact on the result, compared to the alternatives like Z-score and using raw data.

During training, the weights of the neural network were initialized randomly and optimized using Adam~\citep{kingma2014adam}.
The maximum iteration was 5k for all experiments. Early stopping was performed based on the validation loss.



For multiple realizations, a random split is performed once per realization, to prevent over-fitting.
``Realization'' refers to the randomized experiments for each input features $\mathbf{X}$ and treatment $T$.

Unless mentioned otherwise, we run hyperparemeter selection based on the nearest neighbor version of PEHE defined in~\citep{shalit17a}, on the validation set:
${\epsilon_\text{PEHE}}_\text{nn} = \frac{1}{N} \sum_{i=1}^N ((1-2t_i)(y_{j(i)}-y_i) - (\hat{y}^1 - \hat{y}^0) )^2 $
where $j(i)$ is the index of the nearest neighbor to sample $i$ in the opposite treatment group.
This metric is used since we do not have access to true PEHE in real-world settings. 
\vspace{-8pt}




\begin{algorithm}[bt]\small
\KwIn{Dimension for each group of features $D_A,D_B,D_C$; Sample size $N$; Number of realizations $M$.}
\caption{Synthetic Data Generation}
\label{algo:syn}
Compute the weight vector $w$ for each realization $h$ where $\{w_h\}_{h=1}^M \sim \mathcal{U}((0, 0.1)^{D_B+D_C})$\\
\ForEach{sample $(\mathbf{x},t,y^{CF},y^{F},{\mu}^1,{\mu}^0)$} 
{   Compute the mean of A,B,C by
    ${\mu}_A \sim \mathcal{N}(0, 5)$,
    ${\mu}_B \sim \mathcal{N}(4, 2)$,
    $ {\mu}_C \sim \mathcal{N}(6, 2)$\;
    A = $\{{A_{i}}\}_{i=1}^{D_A} \sim \mathcal{N}({\mu}_A, 1)$,
    B = $\{{B_{j}}\}_{j=1}^{D_B}, \sim \mathcal{N}({\mu}_B, 1)$,
    C = $\{{C_{k}}\}_{k=1}^{D_C} \sim \mathcal{N}({\mu}_C, 1) $\;
    $\mathbf{x}$ is generated by concatenation of A,B,C\;
     Generate treatment $t$ by
     $t|(A,B) \sim \text{Bernoulli}(p({t}=1)),$ where $p({t}=1)=1-\text{sigmoid}(0.7*\overline{A}+0.3*\overline{B})$\\
     where $\overline{A}$ and $\overline{B}$ denote the mean of the feature vector A and B respectively;\
     
    \ForEach{each realization $h$ under $\mathbf{x}$}
    {  
    ${\mu}^{0} = w_h^\top \mathbf{x}_{BC}$ where $\mathbf{x}_{BC}$ denote the feature vectors B and C in $\mathbf{x}$ \;
     ${\mu}^{1} =  {\mu}^{0}+10 $\;
        \If{${t}=0$}
        {
             ${y}^{CF}|(B,C) = {\mu}^{1} + \mathcal{N}(0, 1)$; ${y}^{F}|(B,C) = {\mu}^{0} + \mathcal{N}(0, 1)$\;
        }
        \ElseIf{${t}=1$}
        {
             ${y}^{F}|(B,C) = {\mu}^{1} + \mathcal{N}(0, 1)$; ${y}^{CF} |(B,C)= {\mu}^{0} + \mathcal{N}(0, 1)$\;
        }
    }
}
\KwOut{Sample set $\{(\mathbf{x}_i, t_i, {y_i}^{CF}, {y_i}^F, {\mu_i}^1, {\mu_i}^0)\}^N_{i=1}$}
\end{algorithm}

\subsection{Experiment on Synthetic Dataset}
\begin{table}[tbp] 
\setlength{\abovecaptionskip}{0cm}
\setlength{\belowcaptionskip}{0cm}
\centering
\setlength{\tabcolsep}{2.5 pt}
\def\arraystretch{1} 
\caption{Performance comparison on the synthetic dataset over 1000 realizations. The metrics are mean and standard errors of $\sqrt{\epsilon_\text{PEHE}}$, $\sqrt{\epsilon_{\text{PEHEnn}}}$ and $\epsilon_\text{ATE}$. 
Better result with statistical significance by Welch's t-test with $\alpha=0.05$ is highlighted in \textcolor{blue}{blue}.}
\label{tb:syn}
\resizebox{0.9\textwidth}{!}{
  \begin{tabular}{c | ccc | ccc }
    \hline
      \multirow{2}{*}{{\small Methods}} & \multicolumn{3}{c|}{Within-sample} & \multicolumn{3}{c}{Out-of-sample}\\
          & $\sqrt{\epsilon_\text{PEHE}}$ &$\sqrt{\epsilon_\text{PEHEnn}}$ & $\epsilon_\text{ATE}$ & $\sqrt{\epsilon_\text{PEHE}}$ & $\sqrt{\epsilon_\text{PEHEnn}}$ & $\epsilon_\text{ATE}$ \\
    \hline
    CFRW   & 0.258 $\pm$ 0.004 & 1.583 $\pm$ 0.003 & 0.210 $\pm$ 0.008 & 0.257 $\pm$ 0.004 & 1.722 $\pm$ 0.007 & 0.210 $\pm$ 0.008  \\
    RSB(Ours) & \textcolor{blue}{0.237 $\pm$ 0.004} & 1.577 $\pm$ 0.003 & \textcolor{blue}{0.166 $\pm$ 0.006} & \textcolor{blue}{0.237 $\pm$  0.004} & 1.721 $\pm$ 0.007 & \textcolor{blue}{0.167 $\pm$ 0.006} \\
    \hline
  \end{tabular}
}
\vspace{-15pt}
\end{table}

As a sanity check,
we experiment on a synthetic toy dataset to evaluate how well our model handles selection bias in a simple setting, in which the covariates $\mathbf{X}$ is generated by simply concatenating the proposed three types of variables $\mathbf{A}$,$\mathbf{B}$ and $\mathbf{C}$.

The detailed explanation can be found in Algorithm~\ref{algo:syn}.
The feature vectors $\mathbf{A}$,$\mathbf{B}$ and $\mathbf{C}$ are sampled from normal distributions where the variances are fixed but the mean are sampled from another normal distribution for each sample.
%
Since treatment $T$ is binary and only affected by variables $\mathbf{A}$ and $\mathbf{B}$ in our graphical model, a Bernoulli distribution is used to generate $T$ and the probability $p$ is calculated as a $sigmoid$ function applied to a weighted sum of $\mathbf{A}$ and $\mathbf{B}$ to map the value to [0,1].
The noiseless outcome $\mu^0$ are generated by a linear combination of $\mathbf{B}$ and $\mathbf{C}$ where the weights are generated for each realization from a uniform distribution $\mathcal{U}((0, 0.1)^{D_B+D_C})$ where $D_B$ and $D_C$ are the dimensions of $\mathbf{B}$ and $\mathbf{C}$.
$\mu^1$ is then generated by simply adding a constant (10) to $\mu^0$ for all samples.
The noisy outcomes $Y^1$ and $Y^0$ are generated by adding a Gaussian noise $\mathcal{N}(0,1)$ to $\mu^1$ and $\mu^0$ respectively.
The data distribution of the toy dataset is designed to be simple, as the goal is to check if our method of reducing selection bias works in a very simple scenario: all features $\mathbf{ABC}$ are directly observable instead of hidden; the outcomes are linear w.r.t. the features $\mathbf{BC}$ and the ITE is a constant for all samples.

%
We generate 1000 realizations, each contains 1000 samples with 25 covariates (the dimension of $\mathbf{A},\mathbf{B},\mathbf{C}$ are 5,15,5).
We use a 63/27/10 train/valid/test split following the literature.

%

We compare our method RSB with its counterpart CFRW~\citep{shalit17a}, to evaluate how well reducing the selection bias helps in the presence of directly observable confounders.
CFRW is considered as the counterpart since our model RSB follows the same network architecture and can be viewed roughly as CFRW with two additional loss terms: reconstruction loss $L_\text{recon}$ and Pearson correlation coefficient based loss $L_\text{pcc}$. 

The comparison is shown in Table~\ref{tb:syn}. We ran extensive hyperparameter tuning on both methods, using 50 realizations and report the result of the selected best parameter on 1000 realizations.
Although we know the dimensions of the variables $\mathbf{A},\mathbf{B},\mathbf{C}$ when we generate the data, we did not use that information to select the best hyper-parameter. 
Note that the within-sample and out-of-sample results are almost identical on $\sqrt{\epsilon_\text{PEHE}}$ and $\epsilon_\text{ATE}$. This is expected since the true ITE is a constant (10) so if our model produces constant prediction on ITE, the result on training and testing set should be the same. We present the result of $\sqrt{\epsilon_{\text{PEHEnn}}}$ to show that there's indeed a difference between training and testing set.  

The overall result shows that in a dataset generated with selection bias, in a simple setting, reducing the selection bias explicitly using our method helps to improve the counterfactual prediction. 

%

\begin{table}[tbp] 
\setlength{\abovecaptionskip}{0cm}
\setlength{\belowcaptionskip}{0cm}
\centering
\setlength{\tabcolsep}{2.5 pt}
\def\arraystretch{1} 
\caption{Performance comparison on the IHDP dataset over 100 realiazations. The metrics are mean and standard errors of $\sqrt{\epsilon_\text{PEHE}}$ and $\epsilon_\text{ATE}$. Best result with statistical significance by Welch's t-test with $\alpha=0.05$ is highlighted in \textcolor{blue}{blue}. 
Entry '-': not reported in the paper.}
\label{tb:IHDP100}
\resizebox{0.6\textwidth}{!}{
  \begin{tabular}{c | cc | cc }
    \hline
      \multirow{2}{*}{{\small Methods}} & \multicolumn{2}{c|}{Within-sample} & \multicolumn{2}{c}{Out-of-sample}\\
          & $\sqrt{\epsilon_\text{PEHE}}$ & $\epsilon_\text{ATE}$ & $\sqrt{\epsilon_\text{PEHE}}$ & $\epsilon_\text{ATE}$ \\
    \hline
    BNN   & - & - & 2.20 $\pm$ 0.130 & - \\
    Deep-Treat   & - & - & 1.93 $\pm$ 0.070 & - \\
    CFRW   & - & - & 0.88 $\pm$ 0.010  & 0.20 $\pm$ 0.003  \\
    CFR-ISW   & - & - & 0.77 $\pm$ 0.010 & \textcolor{blue}{0.19 $\pm$ 0.003} \\

    \hline
    RSB(Ours) & {0.63 $\pm$ 0.025} & {0.25 +/- 0.033} &\textcolor{blue}{ 0.67 $\pm$ 0.043} & { 0.26 $\pm$ 0.035 } \\
    \hline
  \end{tabular}
}
\end{table}

\begin{table}[tbp] 
\setlength{\abovecaptionskip}{0cm}
\setlength{\belowcaptionskip}{-0.2cm}
\centering
\setlength{\tabcolsep}{2.5 pt}
\def\arraystretch{1} 
  \caption{Performance comparison on the IHDP dataset over 1000 realiazations. The metrics are mean and standard errors of $\sqrt{\epsilon_\text{PEHE}}$ and $\epsilon_\text{ATE}$. Best result with statistical significance by Welch's t-test with $\alpha=0.05$ is highlighted in \textcolor{blue}{blue}.}
  \label{tb:IHDP1000}
\resizebox{0.55\textwidth}{!}{
  \begin{tabular}{c | cc | cc }
    \hline
      \multirow{2}{*}{{\small Methods}} & \multicolumn{2}{c|}{Within-sample} & \multicolumn{2}{c}{Out-of-sample}\\
          & $\sqrt{\epsilon_\text{PEHE}}$ & $\epsilon_\text{ATE}$ & $\sqrt{\epsilon_\text{PEHE}}$ & $\epsilon_\text{ATE}$ \\
    \hline
    k-NN   & 2.1 $\pm$ 0.1 & \textcolor{blue}{0.14 $\pm$ 0.01} & 4.1 $\pm$ 0.2  & 0.79 $\pm$ 0.05  \\
    BART   & 2.1 $\pm$ 0.1 & 0.23 $\pm$ 0.01 & 2.3 $\pm$ 0.1  & 0.34 $\pm$ 0.02  \\
    BNN   & 2.2 $\pm$ 0.1 & 0.37 $\pm$ 0.03 & 2.1 $\pm$ 0.1  & 0.42 $\pm$ 0.03  \\
    TARNET   & 0.88 $\pm$ 0.0 & 0.26 $\pm$ 0.01 & 0.95 $\pm$ 0.0  & \textcolor{blue}{0.28 $\pm$ 0.01}  \\
    CFRW   & 0.71 $\pm$ 0.0 & 0.25 $\pm$ 0.01 & 0.76 $\pm$ 0.0  & \textcolor{blue}{0.27 $\pm$ 0.01}  \\
    CEVAE  & 2.7 $\pm$ 0.1 & 0.25 $\pm$ 0.01 & 2.6 $\pm$ 0.1 & 0.46 $\pm$ 0.02 \\
    \hline
    RSB(Ours) & \textcolor{blue}{0.66 $\pm$ 0.0 } & 0.26 $\pm$ 0.01 & \textcolor{blue}{0.68 $\pm$ 0.0} & \textcolor{blue}{0.27 $\pm$ 0.01 } \\
    \hline
  \end{tabular}
 }
\vspace{-15pt}
\end{table}

\subsection{Benchmark dataset - IHDP}
To further evaluate our method, we benchmark our method in the real-world setting, using
a semi-simulated dataset based on the Infant Health and Development Program (IHDP), introduced by~\citet{hill_bayesian_2011}. The data have features from a real randomized experiment, studying the effect of high-quality childcare and home visits on future cognitive test scores. The IHDP dataset uses a simulated outcome and it also artificially introduces sample selection bias by removing a biased subset of the treated population.
The dataset has 747 samples in total (139 treated and 608 control). 
For each sample, there are multiple realizations of the outcomes corresponding to either of the available treatments \footnote{the counterfactual outcomes are only used for evaluation purposes}. 
We use the same 63/27/10 train/valid/test split as in the literature.

For the comparison with baselines, we test our method under both 100 and 1000 realizations, using the dataset IHDP-100 and IHDP-1000 provided by~\citet{johansson2016learning}. The outcomes in these two datasets are generated with non-linear response surface under setting B in ~\citet{hill_bayesian_2011}.
The results are shown in Table~\ref{tb:IHDP100} and~\ref{tb:IHDP1000}. 
Under 100 realizations, we compare with four other neural network based methods: BNN and Deep-Treat whose results are replicated from~\citep{atan2018deep}, CFRW and CFR-ISW for which the results from~\citep{CFR-ISW} are replicated (results with the hyperparameter selected based on ${\epsilon_\text{PEHE}}_\text{nn}$ for a fair comparison).
Under 1000 realizations, we compare with all baseline methods described in the beginning of Sec.\ref{sec:Experiments}, except for CFR-ISW~\citep{CFR-ISW} and Deep-Treat~\citep{atan2018deep} which only reported results under 100 realizations.

Our method achieves state-of-the-art performance in most metrics.
Since this dataset is simulated from real-world observational data, it is not clear what categories of hidden features are present.
The experimental results show that our method is able to perform well on real-world datasets where the underlying structure of the hidden variables is unknown.

\section{Conclusion}
In this paper, we proposed a novel and intuitive method to reduce selection bias in the problem of estimating the individual treatment effect.
We modeled the input features as generated by three types of latent variables \textbf{A,B,C} (in Figure~\ref{fig:graphic_model}). The variables of type \textbf{A} only cause selection bias while not contributing to the outcome $Y$.
Discarding it would help to alleviate selection bias. 
In order to learn the representation of \textbf{A}, an auto-encoder is used to learn the representations of features. We view the learned representations as two random vectors corresponding to categories \textbf{A} and \textbf{BC}.
We then apply a loss based on Pearson correlation coefficient between any pair of random variables between these two vectors to encourage the linear independence of \textbf{A} and \textbf{BC}.
This allows us to explicitly discard the category (\textbf{A}) that partially induces selection bias and only use the relevant features (\textbf{BC}) for the outcome prediction. 
We tested our approach on both synthetic and simulated real-world tasks, showing that our method achieved state-of-the-art results. 


\bibliographystyle{plainnat}
\bibliography{pgm}

\begin{thebibliography}{12}
\providecommand{\natexlab}[1]{#1}
\providecommand{\url}[1]{\texttt{#1}}
\expandafter\ifx\csname urlstyle\endcsname\relax
  \providecommand{\doi}[1]{doi: #1}\else
  \providecommand{\doi}{doi: \begingroup \urlstyle{rm}\Url}\fi

\bibitem[Atan et~al.(2018)Atan, Jordan, and van~der Schaar]{atan2018deep}
Onur Atan, J~Jordan, and Mihaela van~der Schaar.
\newblock Deep-treat: Learning optimal personalized treatments from
  observational data using neural networks.
\newblock In \emph{AAAI}. AAAI, 2018.

\bibitem[Cheung et~al.(2014)Cheung, Livezey, Bansal, and
  Olshausen]{cheung_discovering_2014}
Brian Cheung, Jesse~A. Livezey, Arjun~K. Bansal, and Bruno~A. Olshausen.
\newblock Discovering {Hidden} {Factors} of {Variation} in {Deep} {Networks}.
\newblock \emph{arXiv:1412.6583 [cs]}, December 2014.
\newblock URL \url{http://arxiv.org/abs/1412.6583}.
\newblock 00075 arXiv: 1412.6583.

\bibitem[Chipman et~al.(2010)Chipman, George, McCulloch,
  et~al.]{chipman2010bart}
Hugh~A Chipman, Edward~I George, Robert~E McCulloch, et~al.
\newblock Bart: Bayesian additive regression trees.
\newblock \emph{The Annals of Applied Statistics}, 4\penalty0 (1):\penalty0
  266--298, 2010.

\bibitem[Hassanpour and Greiner(2019)]{CFR-ISW}
Negar Hassanpour and Russell Greiner.
\newblock Counterfactual regression with importance sampling weights.
\newblock In \emph{Proceedings of the Twenty-Eighth International Joint
  Conference on Artificial Intelligence, {IJCAI-19}}, pages 5880--5887, 7 2019.

\bibitem[Hill(2011)]{hill_bayesian_2011}
Jennifer~L. Hill.
\newblock Bayesian {Nonparametric} {Modeling} for {Causal} {Inference}.
\newblock \emph{Journal of Computational and Graphical Statistics}, 20\penalty0
  (1):\penalty0 217--240, January 2011.
\newblock ISSN 1061-8600, 1537-2715.

\bibitem[Imbens and Rubin(2015)]{imbens2015causal}
Guido~W Imbens and Donald~B Rubin.
\newblock \emph{Causal inference in statistics, social, and biomedical
  sciences}.
\newblock Cambridge University Press, 2015.

\bibitem[Imbens and Wooldridge(2009)]{imbens2009recent}
Guido~W Imbens and Jeffrey~M Wooldridge.
\newblock Recent developments in the econometrics of program evaluation.
\newblock \emph{Journal of economic literature}, 47\penalty0 (1):\penalty0
  5--86, 2009.

\bibitem[Johansson et~al.(2016)Johansson, Shalit, and
  Sontag]{johansson2016learning}
Fredrik~D Johansson, Uri Shalit, and David Sontag.
\newblock Learning {Representations} for {Counterfactual} {Inference}.
\newblock In \emph{International Conference on Machine Learning}, pages
  3020--3029, 2016.

\bibitem[Kingma and Ba(2014)]{kingma2014adam}
Diederik~P Kingma and Jimmy Ba.
\newblock Adam: A method for stochastic optimization.
\newblock \emph{arXiv preprint arXiv:1412.6980}, 2014.

\bibitem[Louizos et~al.(2017)Louizos, Shalit, Mooij, Sontag, Zemel, and
  Welling]{louizos2017causal}
Christos Louizos, Uri Shalit, Joris~M Mooij, David Sontag, Richard Zemel, and
  Max Welling.
\newblock Causal effect inference with deep latent-variable models.
\newblock In \emph{Advances in Neural Information Processing Systems}, pages
  6446--6456, 2017.

\bibitem[Mansour et~al.(2009)Mansour, Mohri, and
  Rostamizadeh]{mansour2009domain}
Yishay Mansour, Mehryar Mohri, and Afshin Rostamizadeh.
\newblock Domain adaptation: Learning bounds and algorithms.
\newblock \emph{arXiv preprint arXiv:0902.3430}, 2009.

\bibitem[Shalit et~al.(2017)Shalit, Johansson, and Sontag]{shalit17a}
Uri Shalit, Fredrik~D. Johansson, and David Sontag.
\newblock Estimating individual treatment effect: generalization bounds and
  algorithms.
\newblock In \emph{International Conference on Machine Learning (ICML)}, pages
  3076--3085, 2017.

\end{thebibliography}

\end{document}